\documentclass[10pt,twocolumn,letterpaper]{article}

\usepackage[pagenumbers]{cvpr} 

\usepackage{graphicx}
\usepackage{amsmath}
\usepackage{amssymb}
\usepackage{booktabs}
\usepackage{arydshln}

\usepackage[hyphens]{url}
\usepackage[pagebackref,breaklinks,colorlinks]{hyperref}
\usepackage[hyphenbreaks]{breakurl}

\usepackage[capitalize]{cleveref}
\crefname{section}{Sec.}{Secs.}
\Crefname{section}{Section}{Sections}
\Crefname{table}{Table}{Tables}
\crefname{table}{Tab.}{Tabs.}

\def\cvprPaperID{*****} 
\def\confName{CVPR}
\def\confYear{2022}

\begin{document}

\title{AVATAR submission to the Ego4D AV Transcription Challenge}

\author{Paul Hongsuck Seo \quad Arsha Nagrani \quad Cordelia Schmid\\
Google Research\\
{\tt\small \{phseo, anagrani, cordelias\}@google.com}
}
\maketitle

\begin{abstract}
In this report, we describe our submission to the Ego4D AudioVisual (AV) Speech Transcription Challenge 2022. Our pipeline is based on AVATAR, a state of the art encoder-decoder model for AV-ASR that performs early fusion of spectrograms and RGB images. We describe the datasets, experimental settings and ablations. Our final method achieves a WER of 68.40 on the challenge test set, outperforming the baseline by 43.7\%, and winning the challenge.
\end{abstract}

\section{Introduction}
\label{sec:intro}
This is the first year of the Ego4D~\cite{grauman2022ego4d} challenges, a set of 16 challenges covering 5 benchmarks. This report describes our submission to the Audio-visual (AV) Automatic Speech Recognition Challenge\footnote{\url{https://eval.ai/web/challenges/challenge-page/1637/overview}}, where the goal is to transcribe the content spoken in a given egocentric video. The output is expected to have timestamps at the word-level.

For the ego-centric domain, the visual stream can
provide strong cues for improving ASR, particularly in cases
where the audio is degraded or corrupted. While most AV-ASR works focus on lip motion~\cite{Noda2014AudiovisualSR,Tamura2015AudiovisualSR,Chung2017LipRS,Afouras2018DeepAS,Petridis2018EndtoEndAS,Makino2019RecurrentNN,Ma2021EndToEndAS,Serdyuk2021AudioVisualSR} (using video crops centered around the speaker’s mouth), this is less commonly available as a cue in the egocentric domain. Hence we use our recently proposed AVATAR~\cite{gabeur2022avatar} model, which ingests full RGB frames from the video, and can therefore use visual cues beyond lip motion that might be useful for the task of ASR, such as the hand movements of a speaker, the presence of certain objects that are being described or even the background location.
Our transcription pipeline consists of audio (and video) segmentation using fixed length chunks, segment transcription using the audio-visual AVATAR model, and 
finally, post-hoc forced alignment to obtain word boundaries.

We show the following; (i) pretraining on the HowTo100M~\cite{miech19howto100m} dataset improves performance on the Ego4D test set, (ii) adding RGB frames to the input and applying word masking techniques improve performance with ground-truth segmentation; (iii) our final model is able to achieve a WER of 68.40 on the challenge test set, outperforming the baseline by 43.7\%. We note that the largest source of error in our pipeline in the fixed length segmentation at the start, which can mask the gains that can be obtained from visual information. This will be the focus for future work and is described in more detail in Section~\ref{sec:discussion}.
In the rest of this report we describe the method, datasets, implementation details and results. 
\section{Method}~\label{sec:method} 
Our method is based on the AVATAR~\cite{gabeur2022avatar} model for audiovisual ASR, which we adapt to the specifics of the Ego4D challenge. In the original paper, AVATAR was trained and evaluted on pre-cropped segments provided by the datasets.  
The Ego4D AV dataset, however, does not provide segment boundaries for the test set. 
Additionally, AVATAR does not produce word timestamps, which are required outputs for the challenge.

In order to tackle the above issues, our transcription pipeline first performs audio (and video) segmentation;
we simply segment the video into fixed length (20 sec) chunks.
Then, a transcription of each segment is predicted using our AVATAR model. 
Finally, word boundaries are obtained in a post-hoc manner using an off-the-shelf forced alignment tool.
This full transcription pipeline allows us to obtain a sequence of timestamped words from a long untrimmed video of arbitrary length.

\subsection{AVATAR model}
\noindent\textbf{Inputs.} Our model is trained on spectrograms and RGB frames extracted from cropped audio segments. We train on 80-dimensional spectrograms from the 16kHz raw speech signal using a Hamming window of 25ms and a stride of 10ms. We randomly extract 2 frames at 1 fps from each input video clip, which are then converted into tokens by extracting $16\times16\times2$ tubelets resulting in a total of $14\times14 = 146$ input tokens (the image resolution is $224\times224$). \\
\noindent\textbf{Architecture.} We follow the configuration outlined in AVATAR~\cite{gabeur2022avatar}. AVATAR consists of a multimodal encoder used to encode both RGB frames and audio spectrograms, and a transformer decoder which produces the natural language speech recognition output. 
The encoder is MBT~\cite{nagrani2021attention}, a transformer based multimodal encoder while the decoder is an auto-regressive transformer decoder~\cite{vaswani2017transformer} consisting of 8 layers and 4 attention heads. \\ 
\noindent\textbf{Post-processing.} For forced-alignment, we use the Montreal Forced Aligner (MFA)~\cite{mfa} with an English pronunciation dictionary~\cite{mfa_english_mfa_dictionary_2022} and a pretrained acoustic model~\cite{mfa_english_mfa_acoustic_2022} provided by the tool.

\section{Experiments}
\subsection{Datasets}
We pretrain our model on the HowTo100M dataset~\cite{miech19howto100m} using pseudo-GT transcriptions obtained from an off-the-shelf ASR model. HowTo100M is a large dataset of instructional videos, where the vision is often correlated with the speech, and where egocentric viewpoints may be more common. We test both random and content word masking strategies during training, as described in \cite{gabeur2022avatar}.
The pretrained model is then finetuned on the Ego4D dataset.

\subsection{Results with Ground Truth Segmentation}
The training and validation set of the Ego4D dataset come with ground-truth segmentation boundaries, where longer clips are broken up into shorter clips based on sentences.
We first test our ASR model alone on the validation set using these groundtruth segments in training and validation.
In Table~\ref{tab:gt_seg_only}, we first observe that pretraining improves performance by a significant margin (over 40\% absolute error reduction).
In fact, the pretrained model without finetuning (row 1 and 2) already outperforms the models that are trained on Ego4D from scratch (row 3 and 4). We also observe that adding vision (A+V) improves the performance over audio only (A).

\subsection{Segmentation Schemes}
Note that while ground-truth segmentation boundaries are provided for the Ego4D training set, they are not provided for the test set. 
We show that training our model with ground-truth segment boundaries, but adopting a fixed length segmentation during inference leads to a drop in performance in Table~\ref{tab:segmentations}.
To close this gap between segmentation schemes at training and inference, we experiment with fixed-length segmentation during training as well, which lowers the error rate.
For this, we first compute word alignments for the examples in the training set using a forced-alignment tool. Then, we segment the input video into chunks of fixed length, and assign each aligned word to its corresponding segment.
We find that the gain of A+V over A is largely hidden when fixed segmentation is used.
This is partly because the models focus more on reducing dominating error cases, which are misrecognized words from the large amount of silence present in the fixed length segments.

\begin{table}[t]
    \centering
    \caption{Ablations with different pretraining (PT) and finetuning (FT) configurations. Note that the models are evaluated with groundtruth segmented clips on the validation set. Modalities: A: Audio only. A+V: Audio+Vision.}
    \label{tab:gt_seg_only}
    \begin{tabular}{ccccc}\hline
        HowTo100M PT & Ego4D FT & Modality & WER \\ \hline
        $\checkmark$ &  & A & 95.25 \\
        $\checkmark$ &  & A+V & 92.03 \\ \hdashline
        & $\checkmark$ & A &  99.37 \\
        & $\checkmark$ & A+V & 100.40 \\\hdashline
        $\checkmark$ & $\checkmark$ & A & 56.83 \\
        $\checkmark$ & $\checkmark$ & A+V & 55.27 \\ \hline
    \end{tabular}
\end{table}

\begin{table}[t]
    \centering
    \caption{
    Ablations with different segmentation schemes during \textit{training} on Ego4D. Models are evaluated with fixed segmentation on the validation set.}
    \label{tab:segmentations}
    \begin{tabular}{ccc}\hline
        Training Data & Modality & WER \\ \hline
        GT segmentation & A & 73.11 \\
        GT segmentation & A+V & 69.91 \\\hdashline
        Fixed segmentation & A & 64.23 \\
        Fixed segmentation & A+V & 64.02 \\ \hline
    \end{tabular}
\end{table}

\begin{table}[t]
    \centering
    \caption{Ablations with different combination of word masking strategies and segmentation schemes on the validation set. Fixed segments represent fixed-length segments.}
    \label{tab:masking}
    \begin{tabular}{ccc}\hline
        & WER with & WER with  \\
        Word masking strategy & GT segments & fixed segments  \\ \hline
        None & 56.30 & 64.02 \\
        Random word masking & 55.43 & 65.94 \\
        Content word masking & 55.27 & 66.26 \\ \hline
    \end{tabular}
\end{table}

\begin{table}[t]
    \centering
    \caption{Final scores with the full pipeline on the validation and test sets.
    The scores on the test set are computed in the challenge leaderboard whereas the score on the validation set is computed locally with a reproduced challenge evaluation procedure.}
    \label{tab:final}
    \begin{tabular}{ccc}\hline
        Method & Validation WER & Test WER \\ \hline
        Baseline & -- & 112.10 \\
        Ours & 71.30 & 68.40 \\ \hline
    \end{tabular}
\end{table}

\subsection{Word Masking Strategies}
We also experiment with the masking strategies introduced in \cite{gabeur2022avatar}. These are introduced to encourage the model to pay attention to the visual stream, by masking certain words in the audio signal.
In Table~\ref{tab:masking}, we show that models trained with word masking outperform those without when tested with ground truth segmentation. 
However, the trend reverses when we apply fixed-length segmentation; in this case the model without any word masking performs the best.
We hypothesise that this is because word masking encourages the model to guess words given a masked input by looking at the visual inputs. 
When applied with fixed-length segmentation (where segments contain a large amount of silence), the model trained with word masking may predict additional words for silence in the input by guessing from the visual context.
Therefore, we use the transcription model trained without masking during training in our final pipeline.

\subsection{Results with Full Pipeline}
Table \ref{tab:final} shows the final results with the full transcription pipeline.
Note that the full evaluation pipeline aligns the predicted words to GT-transcriptions using their timestamps; this results in some performance drop on the validation set ($\sim$7\% increased errors compared to the result in Table~\ref{tab:masking}).
Finally, our method achieves an absolute error reduction of 43.7\% on the test set compared to the baseline provided by the challenge organizers.

\section{Discussion}\label{sec:discussion} 
We note that when using ground truth segmentation, the advantages of both visual information and masking training strategies are clear. However, when we apply fixed length segmentation in training and evaluation, the performance of the overall system drops, and the gains from vision and masking strategies are largely hidden. We hypothesise that this is due to the large amount of silence present in the fixed length segments obtained from long untrimmed videos. Future work will explore better segmentation strategies such as those based on voice activity detection (VAD).

{\small
\bibliographystyle{ieee_fullname}
\bibliography{egbib}
}

\end{document}